\documentclass[sigconf]{acmart}

\usepackage{booktabs} 
\usepackage{color}
\usepackage{multirow}
\usepackage{colortbl}
\usepackage{enumitem,kantlipsum}
\usepackage{diagbox}

\setcopyright{rightsretained}

\acmDOI{10.475/123_4}

\acmISBN{123-4567-24-567/08/06}

\acmConference[UNDER REVIEW IN ACM MM]{20}{April}{2017} 

\acmPrice{15.00}

\begin{document}
\title{Where to Play: Retrieval of Video Segments using Natural-Language Queries}


\author{Sangkuk Lee}
\affiliation{%
  \institution{Seoul National University}
  \streetaddress{1 Gwanak-ro, Gwanak-gu}
  \city{Seoul} 
  \state{Republic of Korea} 
  \postcode{151-742}
}
\email{sangkuklee@snu.ac.kr}

\author{Daesik Kim}
\affiliation{%
  \institution{Seoul National University}
  \streetaddress{1 Gwanak-ro, Gwanak-gu}
  \city{Seoul} 
  \state{Republic of Korea} 
  \postcode{151-742}
}
\email{daesik.kim@snu.ac.kr}

\author{Myunggi Lee}
\affiliation{%
  \institution{Seoul National University}
  \streetaddress{1 Gwanak-ro, Gwanak-gu}
  \city{Seoul} 
  \state{Republic of Korea} 
  \postcode{151-742}
}
\email{myunggi89@snu.ac.kr}

\author{Jihye Hwang}
\affiliation{%
  \institution{Seoul National University}
  \streetaddress{1 Gwanak-ro, Gwanak-gu}
  \city{Seoul} 
  \state{Republic of Korea} 
  \postcode{151-742}
}
\email{hjh881120@snu.ac.kr}

\author{Nojun Kwak}
\affiliation{%
  \institution{Seoul National University}
  \streetaddress{1 Gwanak-ro, Gwanak-gu}
  \city{Seoul} 
  \state{Republic of Korea} 
  \postcode{151-742}
}
\email{nojunk@snu.ac.kr}

\renewcommand{\shortauthors}{S. Lee et al.}
\newcommand{\etal}{\textit{et al}. } 
\newcommand{\ie}{\textit{i}.\textit{e}. } 

\begin{abstract}
In this paper, we propose a new approach for retrieval of video segments using natural language queries. Unlike most previous approaches such as concept-based methods or rule-based structured models, the proposed method uses image captioning model to construct sentential queries for visual information. In detail, our approach exploits multiple captions generated by visual features in each image with `Densecap'. Then, the similarities between captions of adjacent images are calculated, which is used to track semantically similar captions over multiple frames. Besides introducing this novel idea of 'tracking by captioning', the proposed method is one of the first approaches that uses a language generation model learned by neural networks to construct semantic query describing the relations and properties of visual information. To evaluate the effectiveness of our approach, we have created a new evaluation dataset, which contains about 348 segments of scenes in 20 movie-trailers. Through quantitative and qualitative evaluation, we show that our method is effective for retrieval of video segments using natural language queries.

\end{abstract}

%
%




\maketitle

\section{Introduction}

\begin{figure}
\includegraphics[height=2in, width=3.2in]{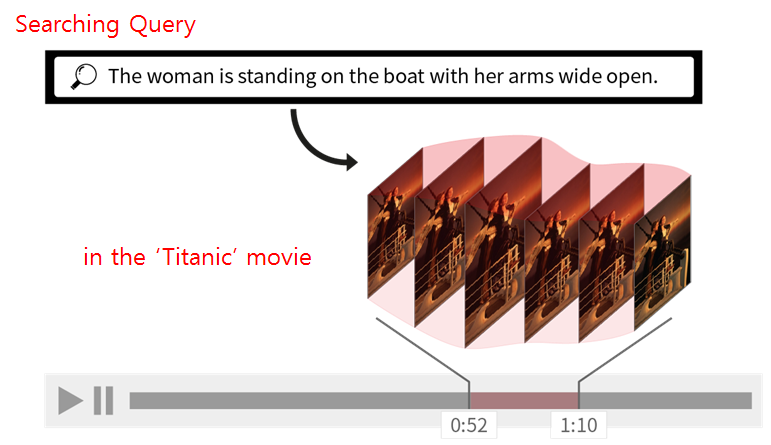}
\caption{An example of retrieving video segments using natural-language queries.}
\label{fig:intro}
\end{figure}

As various video-related services such as YouTube\circledR\ , Facebook\circledR\ and Snapchat\circledR\ have been launched, there has been a rapidly growing interest in technologies of video searches. 

One of the challenges within the field of video searches is to find the time segments of interest given a natural-language query in the form of a sentence or phrase. For example, imagine the following case. One day, you watched the movie `Titanic' in your smart-phone app. A few days later, you wanted to find a scene in  the film that you saw impressively. So you open your app and write down on the search box, like this. ``The woman is standing on the boat with her arms wide open.'' When you hit the button, a few thumbnails of clips appear. Then, you click on the clip you are looking for, and play it again. This process is described in Fig \textcolor{red}{\ref{fig:intro}}. Although most people would sympathize with such simple motif, the problem has actually been technologically challenging to be solved.  

Recent researches related to this are concept-based methods that performs tracking by making use of concepts which are objects detected by object detectors. After finding individual concepts on the tracks, they just took an intersection of them to search scenes corresponding to a specific query \cite{christel2004exploiting, worring2007mediamill, snoek2007learned, tapaswi2014story, aytar2008utilizing}.

Since it is difficult to find semantic relations between concepts, new approaches based on semantic graphs and structured models have been proposed \cite{lin2014visual, barrett2016saying}. To solve the aforementioned problem, they proposed methods of combining models for individual words into a model for an entire sentence. However, because these methods use rule-based structured models or graphs to construct the meaning of a sentence from the meaning of the words in the sentence, it can only deal with queries that fit to the already defined rules. 

It is important to note that most works mentioned above focus on mining characteristic concepts or objects and constructing a sentential query through a rule-based connection between them. On the other hand, in this paper, we propose a new approach which, unlike most previous approaches, uses neural language generation model to construct sentential queries and applies caption-based tracking to search for the scenes corresponding to the queries. This novel idea of `tracking by captioning' will be introduced and highlighted in more detail in later sections. The contributions of this work can be summarized as the following three aspects.

\begin{enumerate}[wide, labelwidth=!, labelindent=0pt]

\item[1)] Rather than constructing the meaning of a sentence from individual information extracted by object or concept detected in an image, we extract sentential queries from visual features in a still image based on a language generation model.  This idea of obtaining sentence from visual features is commonly called as `captioning'. However, since general image captioning generates only one caption for a single image, there is a lack of information for video retrieval. Thus, we use `Densecap'  \cite{johnson2016densecap} here to extract as much captions as possible from a single image. 
This paper is one of the first approaches that uses the language generation model to construct semantic query describing the relations and properties of visual information in videos.

\item[2)] After extracting the captions from all the images in a video with Denscap, tracking is performed by connecting semantically similar captions, rather than connecting objects or concepts. It is a new attempt to find semantically linked segments within a video. We name it as 'tracking by captioning'.

\item[3)] To evaluate performance of our approach, we newly created an evaluation dataset using {\itshape Trailers-Dataset}. 
Through quantitative and qualitative evaluation, we show that our method is effective for retrieval of video segments using natural language queries.

\end{enumerate}

\section{Related Works}

Our proposed approach draws on recent works in semantic search of video segments and image captioning.

\subsection{Semantic Search of Video Segments} 

Most of the recent studies related to this area of semantic search of video segments can be divided into two categories as follows. First, there are concept-based methods that mainly perform tracking by detected objects or concepts. The aim of this line of researches is to match the concepts on the visual tracks. They separately find the video segments related to nouns and verbs, and take the intersection of these set of segments as a visual track \cite{christel2004exploiting, worring2007mediamill, snoek2007learned, tapaswi2014story, aytar2008utilizing}. 
Hu \etal \cite{hu2011survey} surveyed recent works on semantic video search. They note that recent works has focused on detecting nouns and verbs, or using language to search already-existing video annotation. However, the spatial and semantic relations between concepts have rarely been explored. 
Thus, these approaches cannot distinguish two sentences having different meanings consisting of the same words\footnote{For example, it could not distinguish \textit{`the person rode the horse'} versus \textit{`the horse rode the person'} as exemplified in \cite{barrett2016saying}. We will discuss this issue in section \ref{QualEval}}.

Second, there are methods that utilize graphs, or structures to construct complex textual queries \cite{barrett2016saying}, \cite{lin2014visual}.
To solve this problem, they applied the `syntactic trees' describing spatial and semantic relations between concepts.
However, because these methods use rule-based structures or graphs to construct the meaning of a sentence from the meaning of the consisting words, it can only deal with queries that fit to the already defined rules.

It is important to note that most works mentioned above focus on mining characteristic concepts or objects, and constructing a sentential query through a structured connection between them. Unlike these approaches, our approach which uses the neural language generation model can generate sentential queries for videos without the use of graphs, structures, or syntactic trees.

\subsection{Image Captioning} 

Describing images with natural language is one of the primary goals of computer vision. 
To enable this, not only a visual understanding of an image but also an appropriate language model to express the scene in a natural language is needed, which makes the problem more difficult to solve. 
With the tremendous progress in deep neural networks, several methods are proposed for this purpose of image captioning. Among them, one of the successful approaches is the encoder-decoder image captioning framework \cite{vinyals2015show}. 
It encodes an image into a latent representation using a deep convolutional network and decodes the captions through a recurrent neural network. Upon this work, Xu \etal \cite{xu2015show} and Karpathy \etal \cite{karpathy2015deep} developed attention-based neural encoder-decoder networks, respectively. 
They generate each word relevant to spatial images using attention mechanism. 
Along with images, Donahue \etal \cite{donahue2015long} applied long-short term memory (LSTM) to a video to generate the caption of the whole video.

Despite the challenging nature of this task, there has been a recent line of researches attacking the problem of retrieving image and video using natural language queries. Hu \etal \cite{hu2016natural} addressed the task of natural language object retrieval, to localize a target object within a given image based on a natural language query using Spatial Context Recurrent ConvNet. 
Johnson \etal \cite{johnson2016densecap} tackled the problem of generating multiple captions corresponding to a specific region of interest in the name of Densecap. 
They also retrieved the images in the query by multiple captions. 
Podlesnaya \etal  \cite{podlesnaya2016deep} adopted deep features extracted by a convolutional neural network to develop a video retrieval system. 
However, none of the above approaches has tackled the problem of retrieving segments of a video sequence by natural language. 
Our work is deeply based on Densecap \cite{johnson2016densecap}.

\begin{figure*}[th]
\includegraphics[height=2.0in, width=7.0in]{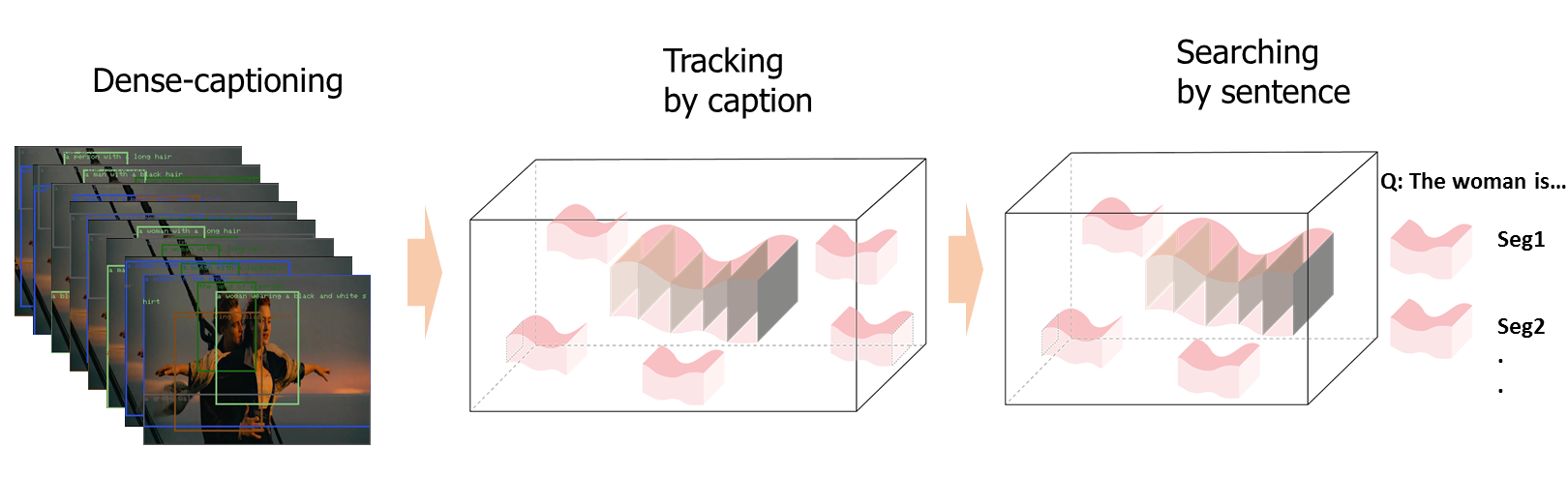}
\caption{Overall structure. (a)extract several boxes with captions using Densecap \cite{johnson2016densecap}  (b)create the tracks that are semantically relevant. (tracking by caption) (c)retrieval of video segments corresponding to the query.}
\label{fig:overview}
\end{figure*}

\section{Method and Architecture}

\subsection{Overview}

The overall structure of our proposed model is illustrated in Fig \textcolor{red}{\ref{fig:overview}}. The model consists of three parts, which work sequentially. We employ a Densecap model \cite{johnson2016densecap} as the first part of our system. 
Densecap was developed with both captioning and object detection methods that could generate captions of detected bounding boxes  for explaining the meanings of regions. 
For each frame, the Densecap model extracts several boxes with captions that explain the circumstances properly. Then, both the box information and the captions generated by a language model are collected.
 
Second, we propose `tracking by caption' method to obtain tracklets which consist of reliable box sequences. In contrast to the conventional `tracking by detection' methods, it focuses not only on positions of boxes but also on semantic meanings of regions derived from the previous parts. As frame sequences pass, we first match positions of boxes with previous frames. Then, for matched boxes, we obtain the similarity between captions and compare the meanings of the regions.
This part can be modelled by several methods that can calculate similarities in natural language. Details are given in section \ref{Similarity}. 
As a result, these similarities are used to connect the boxes in consecutive images to create the tracks that are semantically relevant.

The last part of our model is developed for retrieval of video segments based on the information generated in the previous two parts. After the second part operates, we can get several `semantic tracks', each of which contains frame information and representative caption as the meaning of the track. When a user asks to find segments of a video with a natural language query, the model calculate the similarity between the input query and the representative captions of all the tracks in the video, and propose the tracks that are semantically relevant.

\subsection{Tracking by caption}
\label{Trackingbycaption}

\begin{figure}[bp]
\includegraphics[height=1.7in, width=3.4in]{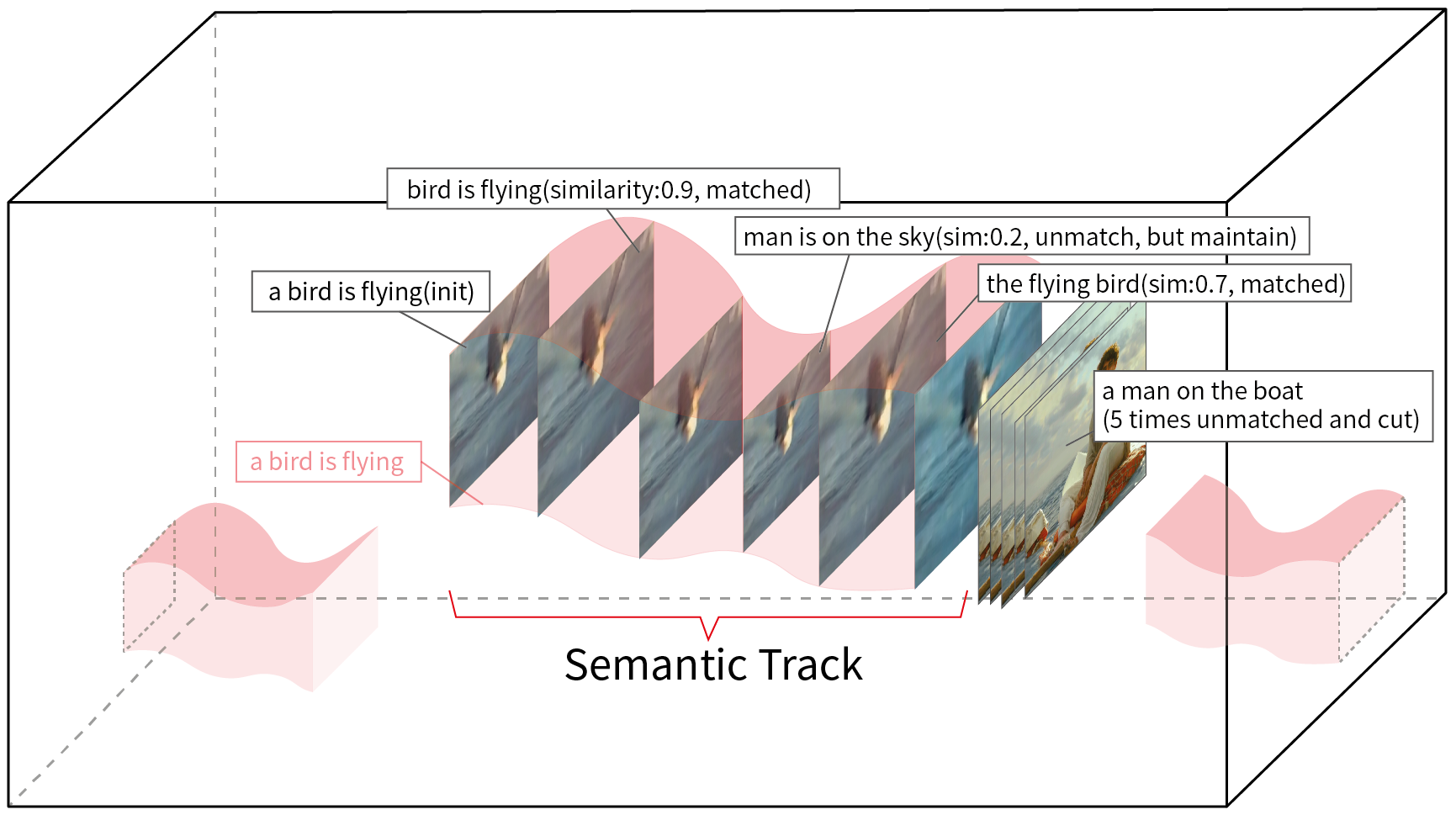}
\caption{Tracking by caption}
\label{fig:tracking_by_caption}
\end{figure}

Tracking by caption is a novel idea of our model. It is a methodology that uses semantic relevance of captions in a way of tracking captions, rather than tracking objects or concepts. 
The overview is illustrated in Fig \ref{fig:tracking_by_caption}. 
This methodology consists of the following three steps.\\
\textbf{Initiation}. Suppose that Densecap generates $N$ boxes with captions for each images. When the first image comes into the model, $N$ boxes are generated, each of which is registered as a new track immediately. When the next image comes up, we find and match the boxes with captions that are semantically similar to the existing tracks. If $M$ boxes are matched, the remaining $N-M$ new boxes are again registered as new tracks. If there are no deleted tracks in this frame, the total number of tracks currently is $2N-M$. The caption representing each track is the caption of the first box registered as the track. While the frame sequences pass, this rule applies to all the frames.\\
\textbf{Matching and Maintenance}. When the similarity between the caption of the track up to the previous frame and the caption of the new box of the current frame is calculated, they are matched only when the similarity is above a certain threshold value. We name this threshold as a `track similarity threshold'. Based on this `track similarity threshold', boxes with semantic relevance are linked to form one track. We call these tracks as `semantic tracks'. If there is no appropriate box to match near a track, the track retains the previous information. All of this is illustrated in Fig \ref{fig:tracking_by_caption}.\\
\textbf{Cutting and Storing}. Basically, when the scene ends or the semantic similarity drops significantly, the update of the corresponding semantic track is stopped. This is named as `cutting'. More specifically, if the number of frames a similar box does not come out exceeds a predetermined threshold, the corresponding track is cut off. In this work, we set the `cutting threshold' to 5. These tracks, by themselves, contain information on the corresponding video segments that started and ended within a certain time span, and are collected and stored in memory or a file.

\subsection{Calculating Similarity}
\label{Similarity}

In both processes of tracking by captions and searching by queries, one of the important things is comparing corresponding captions to decide whether they are matched or not. In this work, we employ a couple of popular methods for embedding queries and captions to sentence vectors, \ie, {\itshape Word2Vec (w2v)} \cite{mikolov2013distributed} and {\itshape Skip-thoughts vector (stv)} \cite{kiros2015skip}. 
%

The first approach in our work makes use of the Word2Vec, which is one of the popular methods to embed words to vector representations. By Word2Vec, it is possible to convert each word in a sentence to a word vector from a pretrained embedding matrix. 
Then we average the corresponding word vectors for each sentence along the word dimension to obtain one vector per sentence. 
However, an average of word vectors is not likely to represent the meaning of a sentence. 
Therefore, as a second approach, we incorporate the skip-thought model to get a vector from a sentence at once. 
The skip-thought model is similar to word2vec model, but it handles sentence-level representations. Therefore, a skip-thought encoder can be used as a generic sentence vector extractor. 

Since both methods have their own characteristics, we conducted experiments to compare the two methods on several conditions.
After extracting sentence vector using Word2Vec or Skip-thought, we use the cosine similarity metric between sentence vectors as a similarity measure. 

\section{Experiments}

\subsection{A New Dataset for Evaluation}
\label{sec:dataset}

There are publicly available datasets for the task of video segments search in current challenges, such as {\itshape ActivityNet} \cite{caba2015activitynet} and {\itshape TRECVID} \cite{2016trecvidawad}.
Especially, these are the `activity detection' task in {\itshape ActivityNet (2016)}, and the `Ad-Voc Search (AVS)' task in {\itshape TRECVID (2016)}.

However, the `activity detection' task in {\itshape ActivityNet (2016)} is in fact a classification problem about pre-determined textual queries. These queries are in the form of a sentence, but it is no different than the class numbers. Since it needs only to find the segments that match the pre-determined query, this task is not appropriate for evaluating our method of natural-language-based video segment search.

The `Ad-Voc Search (AVS)' task in {\itshape TRECVID (2016)} is to find segments of a video containing persons, objects, activities, locations, etc., and the combinations of the former. Also, the task is performed in the `no annotation' condition. That is, there is no annotations in the form of a natural-language describing the video segment.
In this task, the concept information, such as persons, objects, activities, locations, etc., is given first, and then the final evaluation is performed with a sentential query in which the given concept information is randomly combined. 
This task is similar to our work in that it searches for unspecified queries.
However, it is different from our work because it only deals with sentences in the form of a combination of given concepts.

To evaluate performance of our approach, we newly created an evaluation set using the {\itshape Trailers-Dataset}.\footnote{It consists of 474 YouTube official trailers of American movies released in 2010-2014. \textit{https://github.com/tadarsh/movie-trailers-dataset}} 
For the randomly chosen 20 clips from the Trailers-Dataset, we labeled the ground-truth time-stamp (start-frame and end-frame) about a particular scene. 
Since movie trailers are all different and we cannot apply the same queries to different movie trailers, we have chosen a set of queries for each trailer through the following process: 

\begin{enumerate}[wide, labelwidth=!, labelindent=0pt]

\item[i)] We first extracted five bounding boxes and their captions from randomly selected 200 images in the video through Densecap. 

\item[ii)] Next, we ranked the frequency of the extracted captions, and then the top 100 are selected.

\item[iii)] The annotator then freely selects 5 captions out of the 100 selected frequent captions to use as the queries.

\item[iv)] Finally, the annotators find the corresponding segments in each video for the selected queries, and then record the beginning and the ending frames of the segments found. 

\end{enumerate}

This is used as the ground truth of the evaluation.
In this way, we collected the dataset for evaluation, which includes about 348 segments of scenes for 100 queries in 20 movie-trailers. 

It is important to note that this is not the `video-annotation' such as objects, concepts, or, other spatial information, and the combinations of the former. This set is only for evaluation, not for learning. 
It should also be noted that our work ultimately assumes a searchable problem for any query, but it cannot find queries that the language model has not learned. Thus, through the above  process of query generation, we provide queries that are searchable in the video.

In our evaluation process, there was not any pre-determined queries and learnable information about the video segments. This is the `no annotation' condition for video like the AVS task in {\itshape TRECVID (2016)}.
In our method, annotation is only needed for the language model in Densecap to learn captions about still images.

\subsection{Quantitative evaluation}

In this part, we consider the application that retrieves the relevant video segments given a natural-language query using the new dataset described in section \ref{sec:dataset}. First, we performed tracking for all the videos and created semantic tracks. The cosine similarity was used throughout the paper. The similarity threshold in constructing semantic tracks (track similarity threshold) was varied from 0.6 to 0.8. The cutting threshold was set to 5 frames, and the minimum track size was also set to 5 frames, \ie only tracks with length greater than or equal to 5 frames were retained as valid semantic tracks. Then, we searched for the semantic tracks corresponding to a given query as the following:

\begin{enumerate}[wide, labelwidth=!, labelindent=0pt]

\item[i)] For each input query, a set of tracks with similarity higher than the predetermined threshold value is proposed by the algorithm. We name this threshold as the `search similarity threshold', which is used to find tracks similar to the query entered in the search phase. Note that it is different from the `track similarity threshold' described in section \ref{Trackingbycaption} which is used in constructing semantic tracks.

\item[ii)] The IoU (intersection over union) between the proposed set of tracks and the ground truth is calculated by comparing the beginning and ending frames between the proposals and the ground truth.

\item[iii)] The performance of the proposed method is measured with recall, precision, and mAP.

\end{enumerate}

%
\begin{figure}[tbp]
\includegraphics[width=\linewidth]{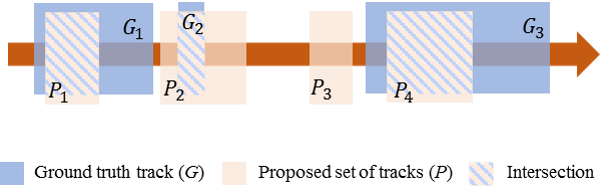}
\caption{The concept of IoU, Recall, and Precision in frames of video segments. If the IoU threshold is set to 0.3, $P_1$ and $P_4$ which have IoU of 0.4 and 0.5 with $G_1$ and $G_3$ respectively are considered as correct detection, while $P_2$ and $P_3$ are considered as false alarms because IoU of $P_2$ with $G_2$ is less than 0.3 and $P_3$ has no overlapping ground truth. Therefore, for this example, the recall and the precision are 66\% and 50\%, respectively.}
\label{fig:IOU}
\end{figure}

Note that, unlike the general definition of IoU used in object detection, here we have defined IoU based on the frames of ground truth segments and proposed tracks. Also, recall and precision are slightly different from the general ones. 
Consider there are $N_t$ ground truth segments $\{G_i | i \in \{1, \cdots, N_t\} \}$ in a video for a specific query. (These are annotated by humans.) For this query, assume that the proposed method outputs $N_p$ proposal semantic tracks $\{P_i | i \in \{1, \cdots, N_p\} \}$. For each pair of ground truth and proposal ($G_i, P_j$), we compute the IoU value and if it is greater than the IoU threshold, $G_i$ is considered to be detected and $P_j$ is marked as a good proposal. After computing IoU values for all the pairs, we can count the number of detected tracks $N_d$ and the number of good proposals $N_g$\footnote{Note that $N_d$ and $N_g$ can be different for a small IoU threshold.}.  Then, the precision is computed as $N_g/N_p$, while the recall is calculated as $N_d/N_t$. See the illustration in Fig \textcolor{red}{\ref{fig:IOU}}.

We calculated the recall and precision by varying the IoU threshold from 0.1 to 0.9. 
Note that the `search similarity threshold' $S_{sim}$ can be used to adjust the recall for a given IoU threshold, \ie if $S_{sim}$ is low, many semantic tracks are proposed and the recall tends to increase. 
This way, the average precision (AP) can be calculated by taking the average of precisions at different recall values in $\{0.0, 0.1, \cdots, 1.0\}$.  
The mean AP (mAP) is obtained by taking the mean of AP for all the input queries.


For all the experiments, we compared the performance of word2vec and skip-thought vector as a model for embedding a sentence into a vector and measuring the similarity.

Table \textcolor{red}{\ref{tab:tab_1}} shows the performance (precision and recall) comparison of the sentence embedding schemes with different track and search similarity thresholds, which are abbreviated as $T_{sim}$ and $S_{sim}$, respectively, in the table. The recall and the precision are computed based on the number of proposed tracks that have IoU with ground truth exceeding the IoU threshold of 0.3.

\begin{table}[tbp]
  \centering
    \small
        \setlength\tabcolsep{2pt}
  \caption{Recalls and precisions for different track similarity thresholds ($T_{sim}$) and search similarity thresholds ($S_{sim}$). IoU threshold is fixed to 0.3}
    \begin{tabular}{c|c|ccc|ccc}
    \toprule
    \multicolumn{1}{r}{} &       & \multicolumn{3}{c|}{stv} & \multicolumn{3}{c}{w2v} \\
    \midrule
    \multicolumn{1}{r}{} &\diagbox[width=3em]{ $S_{sim}$}{ $T_{sim}$}  & 0.6   & 0.7   & 0.8   & 0.6   & 0.7   & 0.8 \\
    \midrule
    \midrule
    \multirow{3}[2]{*}{Recall} & 0.6   & \cellcolor[rgb]{ .651,  .651,  .651} 0.783 & \cellcolor[rgb]{ .749,  .749,  .749} 0.606 & \cellcolor[rgb]{ .851,  .851,  .851} 0.422 & \cellcolor[rgb]{ .651,  .651,  .651} 0.691 & \cellcolor[rgb]{ .749,  .749,  .749} 0.542 & \cellcolor[rgb]{ .851,  .851,  .851} 0.428 \\
          & 0.7   & \cellcolor[rgb]{ .749,  .749,  .749} 0.711 & \cellcolor[rgb]{ .851,  .851,  .851} 0.527 & \cellcolor[rgb]{ .949,  .949,  .949} 0.356 & \cellcolor[rgb]{ .749,  .749,  .749} 0.598 & \cellcolor[rgb]{ .851,  .851,  .851} 0.477 & \cellcolor[rgb]{ .949,  .949,  .949} 0.37 \\
          & 0.8   & \cellcolor[rgb]{ .851,  .851,  .851} 0.659 & \cellcolor[rgb]{ .949,  .949,  .949} 0.471 & \cellcolor[rgb]{ .949,  .949,  .949} 0.336 & \cellcolor[rgb]{ .749,  .749,  .749} 0.561 & \cellcolor[rgb]{ .949,  .949,  .949} 0.444 & 0.376 \\
    \midrule
    \multirow{3}[2]{*}{Precision} & 0.6   & \cellcolor[rgb]{ .851,  .851,  .851} 0.092 & \cellcolor[rgb]{ .749,  .749,  .749} 0.156 & \cellcolor[rgb]{ .651,  .651,  .651} 0.194 & \cellcolor[rgb]{ .851,  .851,  .851} 0.171 & \cellcolor[rgb]{ .749,  .749,  .749} 0.211 & \cellcolor[rgb]{ .651,  .651,  .651} 0.273 \\
          & 0.7   & \cellcolor[rgb]{ .949,  .949,  .949} 0.087 & \cellcolor[rgb]{ .851,  .851,  .851} 0.146 & \cellcolor[rgb]{ .749,  .749,  .749} 0.168 & \cellcolor[rgb]{ .851,  .851,  .851} 0.161 & \cellcolor[rgb]{ .949,  .949,  .949} 0.191 & \cellcolor[rgb]{ .749,  .749,  .749} 0.23 \\
          & 0.8   & 0.079 & \cellcolor[rgb]{ .949,  .949,  .949} 0.138 & \cellcolor[rgb]{ .851,  .851,  .851} 0.15 & 0.144 & \cellcolor[rgb]{ .851,  .851,  .851} 0.175 & \cellcolor[rgb]{ .749,  .749,  .749} 0.228 \\
    \bottomrule
    \end{tabular}%
  \label{tab:tab_1}%
\end{table}%

Track similarity threshold is used to construct semantic tracks in the tracking phase. As this value increases, the connection between consecutive image captions tends to decrease and the number of tracks could be reduced. 
In general, this reduces recall, since it essentially reduces the number of tracks that can be offered, regardless of the IoU threshold.
On the contrary, as this value decreases, false positive tracks could be increased, because many tracks having relatively weak semantic similarity are generated. This reduces the precision in general. Table \textcolor{red}{\ref{tab:tab_1}} shows this trend without exception. 

On the other hand, search similarity threshold is used to find tracks similar to the query entered in the search phase. As this value increases, the proposal for tracks that are semantically similar to the input query is reduced. Thus, the recall tends to be reduced although there is an exception for word2vec at $T_{sim} = 0.8$. In this case, as $S_{sim}$ is increased from 0.7 to 0.8, the recall increases slightly from 0.37 to 0.376.
It is noted that increasing search similarity threshold does not increase precision, since the search similarity threshold only controls the selection among the existing semantic tracks, not affects the creation of new tracks. On the contrary, in our experiment, as search similarity threshold increases, the precision decreases slightly for all the cases. 

Overall, model with skip-thought vector shows better performance in recall and word2vec has better performance in precision. As described in Section \ref{Similarity}, word2vec is basically a word vector extractor. Therefore, average of word vectors does not fully express the meaning of the sentence. On the other hand, skip-thought vector has a sentence-level representation that converts a sentence directly into a vector. Since it is likely to robust to order of word and grammar, more tracks could be connected at tracking phase, and more similar tracks could be proposed at search phase. As a result, the performance of recall increases, and the performance of precision decreases.


\begin{figure}[tbp]
\includegraphics[height=2.0in, width=3.0in]{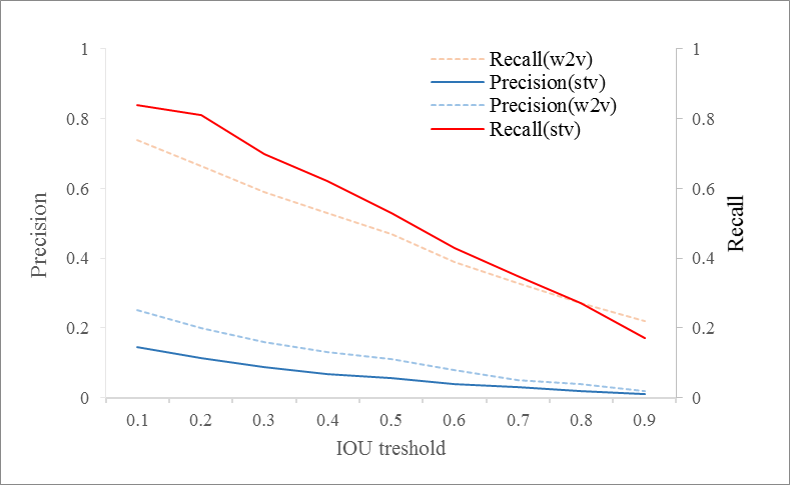}
\caption{Recall and precision according to changes in IoU threshold. $T_{sim}$ and $S_{sim}$ are fixed to 0.7 and 0.6, respectively.}
\label{fig:recall-iou}
\end{figure}

Fig \textcolor{red}{\ref{fig:recall-iou}} shows the change in recall and precision as the IoU threshold changes. $T_{sim}$ is fix at 0.7 and $S_{sim}$ is fixed at 0.6. As in Table \textcolor{red}{\ref{tab:tab_1}}, skip-thought vector has better performance in recall and word2vec has better performance in precision. 

\newcolumntype{P}[1]{>{\centering\arraybackslash}p{#1}}
\begin{table}[t]
  \centering
  \scriptsize 
  \caption{Performance (mAP) comparison of skip-thought vector (stv) and word2vec (w2v)}
      \begin{center}

    \resizebox{\columnwidth}{!}{
    \begin{tabular}{P{2.5cm}|P{0.8cm}P{0.8cm}}
    \toprule
          & w2v   & stv \\
    \midrule
    mAP   & 0.549 & 0.654 \\
    mAP ($Recall \geq 0.5$) & 0.259 & 0.323 \\
    \bottomrule
    \end{tabular}
    }%
        \end{center}

  \label{tab:tab_2}%
\end{table}%

Table \textcolor{red}{\ref{tab:tab_2}} shows the mAP of skip-thought vector and word2vec. In order to prevent the high precision values at recall less than 0.5 from predominantly affecting the performance of the mAP, we further calculated the mAP for a case where recall is 0.5 or more. 
In the table, we can see that the mAP of skip-thought vector is better than word2vec.

\subsection{Qualitative Evaluation}
\label{QualEval}

For qualitative performance evaluation, we have created a simple demo for retrieval of video segments that can be found in the supplementary video. In this section, we will cover some interesting issues and we show some proposal examples for segments that are semantically similar to the input queries. 

\textbf{Distinguishing two sentences with different meanings consisting of the same words}. 
This problem was raised in \cite{barrett2016saying}. Since conventional methods basically take an object or concept-based approach, inevitably they have to use graphs, structures, or syntactic trees.
On the other hand, since our method is a sentence-based approach, we can solve this problem using semantic similarity, without the use of graphs, structures, or syntactic trees.
For the two sentences \textit{`the person rode the horse'} and \textit{`the horse rode the person'}, the cosine similarity calculated using skip-thought vector is 0.68 which is smaller than 1. Therefore, the application can prevent the two sentences from being connected or retrieved through the threshold setting. 

\begin{figure*}[tbp]
\includegraphics[width=\textwidth]{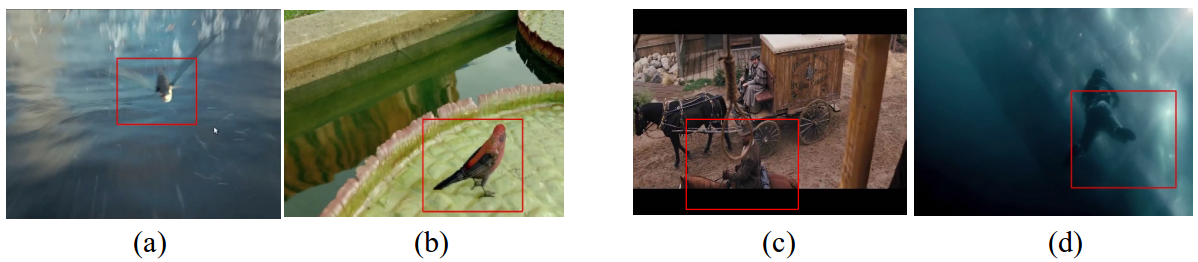}
\caption{Retrieval of video segments about different states and behaviors of the same object. (a)`the bird is flying', (b)`a bird on the branches', and (c)`a man riding a horse', (d)`a man in the water'. }
\label{fig:qual1}
\end{figure*}

\textbf{Searching different states and behaviors of the same object}. Our task is not just to find simple objects, but to be able to understand complex states and behaviors. For example, two sentences such as `The bird is flying' and `a bird on the branches' could be searchable separately for each query. These are shown in Figure \textcolor{red}{\ref{fig:qual1}}.

\begin{figure*}[tbp]
\includegraphics[width=\textwidth]{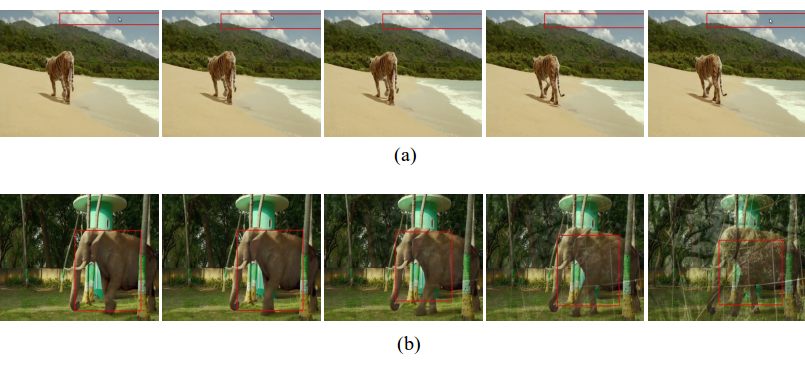}
\caption{Retrieval of video segments and localization. Here, we show 5 consecutive frames with bounding box that corresponds to each query. (a) The detection box is located in the `the cloudy blue sky' at the upper right of the image. (b) the detection box is located on `the elephant on the grass' in the center of the image}
\label{fig:qual2}
\end{figure*}

\textbf{Searching and localization}. Our method is based on the Densecap, which combines image captioning and object detection. Basically, the Denscap looks for an area with objects or visual information, and generates a description that contains semantic information about the area. In our method, even when performing `tracking by caption', tracking is performed based on area information with objects or visual information.
Therefore, our application not only finds the video segments corresponding to the input query, but also knows where the information is located in the image within the video. Specifically, it is represented as a boxes in the image of each frame. These are shown in Figure \textcolor{red}{\ref{fig:qual2}}.

\begin{figure*}[tbp]
\includegraphics[height=3.0in, width=6.0in]{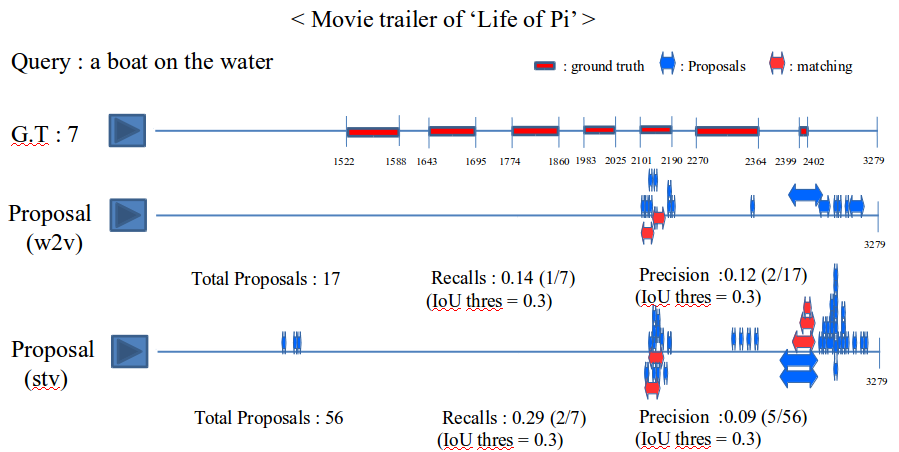}
\caption{Performance of the proposed method on a real movie trailer.}
\label{fig:qual3}
\end{figure*}

Finally, Figure \textcolor{red}{\ref{fig:qual3}} shows the result of the proposed method on a movie trailer. Note that skip-thought vector proposed many proposals such that more than 2 proposal tracks hit one ground truth and the number of detected tracks ($N_d = 2$) and the number of good proposals ($N_g = 5$) are different.

\section{Conclusion}

In this paper, we proposed a novel approach for searching segments of videos from natural-language queries.  We build a pipeline which exploits the Densecap model and the proposed tracking method. Most of all, we developed the `tracking by caption' method which uses semantic relevance of captions in a way of tracking captions, rather than tracking objects or concepts. After tracking is completed, the model extracts several semantic tracks which represent spatio-temporal segments in a video. Then the model is able to offer matched semantic tracks which users need to search. Our proposed method also shows significant flexibilities when a user try to find scenes in a movie. It only necessitate describing a scene with natural language queries that are used in real life. 

Moreover, we created a new evaluation dataset to evaluate the performances of the proposed video segment search method quantitatively. With the dataset, experimental results show that our proposed model could be applied in practice meaningfully. In the future, we plan to develop our model as an end-to-end model to avoid accumulated errors that often occur in a pipeline.

%

%
%

%


\end{document}